# Determining the best classifier for predicting the value of a Boolean field on a blood donor database

Ritabrata Maiti[1]

[1]Delhi Technological University

**Abstract**

**Motivation:** Thanks to digitization, we often have access to large databases, consisting of various fields of information, ranging from numbers to texts and even boolean values.
Such databases lend themselves especially well to machine learning, classification and big data analysis tasks. We are able to train classifiers, using already existing data and use them for predicting the values of a certain field, given that we have information regarding the other fields.

Most specifically, in this study, we look at the Electronic Health Records (EHRs) that are compiled by hospitals. These EHRs are convenient means of accessing data of individual patients, but there processing as a whole still remains a task. However, EHRs that are composed of coherent, well-tabulated structures lend themselves quite well to the application to machine language, via the usage of classifiers. In this study, we look at a Blood Transfusion Service Center Data Set (Data taken from the Blood Transfusion Service Center in Hsin-Chu City in Taiwan). We used scikit-learn machine learning in python. From Support Vector Machines(SVM), we use Support Vector Classification(SVC), from the linear model we import Perceptron. We also used the K.neighborsclassifier and the decision tree classifiers. Furthermore, we use the TPOT library to find an optimized pipeline using genetic algorithms. Using the above classifiers, we score each one of them using k fold cross-validation.

**Results:** The test program relies on the individual testing of the classifiers. It counts the number of predictions that much the actual value and displays these counts. Using the counts, we are able to decide the best classifier for the given blood donor database. Using the most accurate models, or a collection of these models, we will be able to determine the most accurate prediction for each patient. Here, we wish to determine whether a patient had donated blood in March 2017. This prediction is a boolean value(1 or 0), where 1 denotes that the patient had donated blood and 0 denotes otherwise.

**Contact:** ritabratamaiti@hiretrex.com

**GitHub:** https://github.com/ritabratamaiti/Blooddonorprediction

# 1. Introduction

In the last 20 years, the storage capabilities of electronic media have increased exponentially. As a result, the volume of medical data stored in electronic media has increased exponentially. The various medical data available to us range from images and text to video and audio. These are one of the few data-types that are available and manipulated in medical centers. Often the required data is exploited and analyzed on an individual level. For example, a Magnetic Resonance Imaging(MRI) and a textual health record will be analyzed to establish the diagnosis or disease evolution of a patient.

Among all the different data types, only structured data can be readily used for training machine learning algorithms. This is because tabulated data lends itself especially well to the training of machine language classifiers. Via the classifiers, we are able to work on a dataset, and predict the values present in one field of the given set, provided we have information regarding the other fields of the data set. These machine learning algorithms function by sensing the patterns present in an existing dataset and using those patterns to predict values of missing fields.

However, the problem arises that most of the records in medical databases are unstructured free texts, such as patient health charts and electronic patient reports. These utilize natural languages, i.e languages easily understood and processed by humans but less so by machines. Examples of these include notes on a patients diagnosis, a prescription or even a notes regarding a tissue sample. While the free text can be made machine understandable using NLP algorithms, the most effective pattern reconstruction and predictions in medicine are made using structured databases and classifier algorithms.

The technical core of the program relies on the fact that different training algorithms have different success rates when trained with the same dataset. As a result, we will be able to determine which classifier works best with a given algorithm.

While a doctor's diagnosis remains as invaluable in medicine, with data science, disease diagnosis and prediction becomes far more reliable and efficient. Using algorithms as helpers in disease prediction, early detection and analysis not only assists doctors in disease analysis but also helps in decreasing patient mortality.

In this problem we will be classifying whether a donor has donated blood at a particular time (in this case the March of 2007), on the basis of the parameters:
- The recency of a previous blood donation (number of months since last donation)
- Frequency of blood donation
- Amount of blood donated in cubic centimetres (c.c.)
- Time in months since first donation)

We will be also be using a genetic algorithm to perform hyperparameter tuning on a classifier, and compare this classifier's accuracy score against the default classifiers.

## 2. Methods

Classification can be thought of as two separate problems – binary classification and multiclass classification. In binary classification, a better-understood task, only two classes are involved, whereas multiclass classification involves assigning an object to one of several classes.

In binary classification, we group an object into one of two classes while in multiclass classification we group an object into one of many classes. Our specific problem requires us to decide whether a specific patient had donated blood on a previous date, and as a result, our classifier provides us with a boolean output of either 1 or 0, wherein 1 denotes that the patient had donated blood and 0 denotes otherwise.

Thus, we have to solve a binary classification problem. In order to find out the most efficient classifier, we should train each classifier by the same dataset. Then, we should predict the values 1 or 0, for a dataset of new patients, and determine whether the predictions match.

By finding out the number of matched predictions, we can calculate the percentage matches of each database and subsequently determine the most efficient classifier.

In machine learning and statistics, a classification is a problem of identifying to which of a set of categories (sub- populations) a new observation belongs, on the basis of a training set of data containing observations (or instances) whose category membership is known. An example would be assigning a given email into "spam" or "non-spam" classes or assigning a diagnosis to a given patient as described by observed characteristics of the patient (gender, blood pressure, presence or absence of certain symptoms, etc.). Classification is an example of pattern recognition.

In the terminology of machine learning,[1] classification is considered an instance of supervised learning, i.e. learning where a training set of correctly identified observations is available. The corresponding unsupervised procedure is known as clustering and involves grouping data into categories based on some measure of inherent similarity or distance.

Often, the individual observations are analyzed into a set of quantifiable properties, known variously as explanatory variables or features. These properties may variously be categorical (e.g. "A", "B", "AB" or "O", for blood type), ordinal (e.g. "large", "medium" or "small"), integer-valued (e.g. the number of occurrences of a particular word in an email) or real-valued (e.g. a measurement of blood pressure). Other classifiers work by comparing observations to previous observations by means of a similarity or distance function.

An algorithm that implements classification, especially in a concrete implementation, is known as a classifier. The term "classifier" sometimes also refers to the mathematical function, implemented by a classification algorithm, that maps input data to a category.

Terminology across fields is quite varied. In statistics, where classification is often done with logistic regression or a similar procedure, the properties of observations are termed explanatory variables (or independent variables, regressors, etc.), and the categories to be predicted are known as outcomes, which are considered to be possible values of the dependent variable. In machine learning, the observations are often known as instances, the explanatory variables are termed features (grouped into a feature vector), and the possible categories to be predicted are classes. Other fields may use different terminology: e.g. in community ecology, the term "classification" normally refers to cluster analysis, i.e. a type of unsupervised learning, rather than the supervised learning described in this article.

## 2.1 Classifier Selection

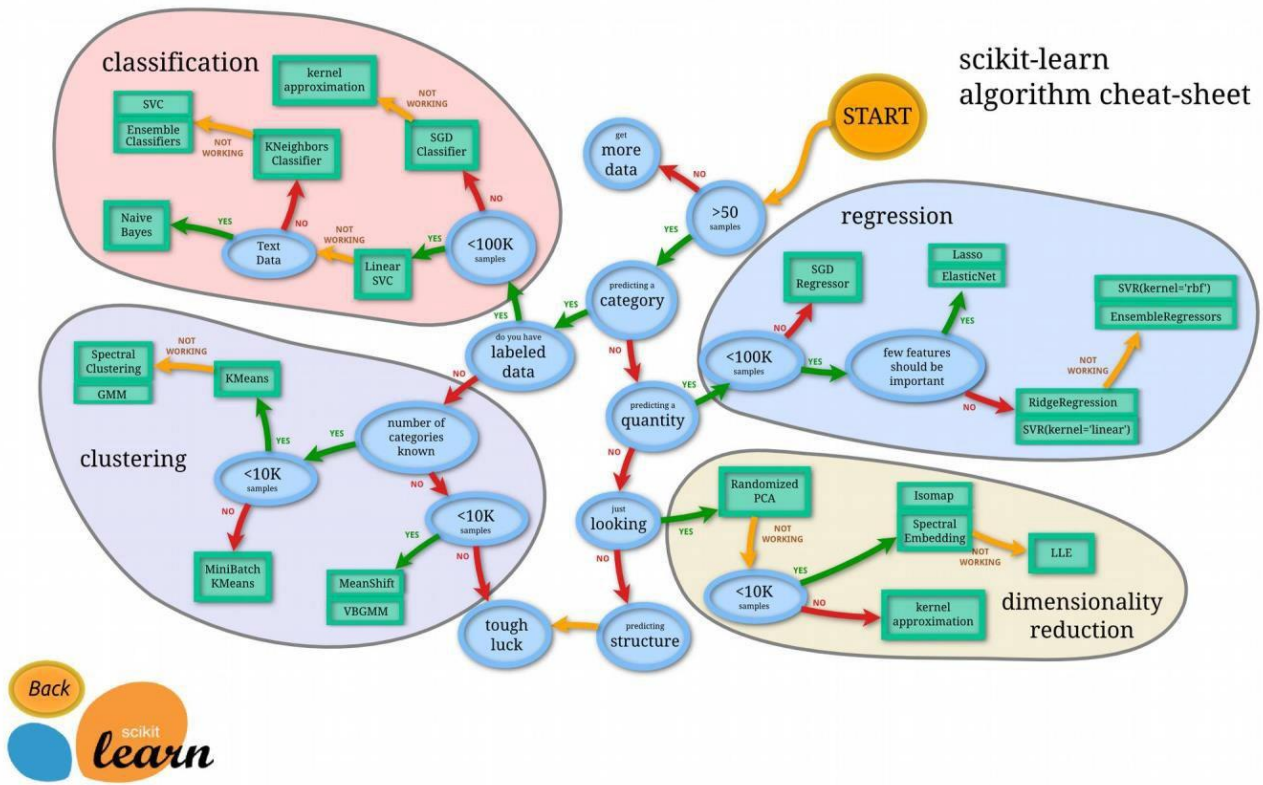

Via usage of the classifier selection diagram given here: We have subsequently followed the steps:

- >50 samples : Yes
- Predicting a category : Yes
- Labelled Data : Yes
- Therefore classification problem
- <100k samples : Yes

Thus we have chosen the following classifiers:
1. SVC
2. Perceptron
3. KneighborsClassifier
4. Decision Tree Classifier

Furthermore, we include a Naïve Bayes Classifier in our tests, because of its high accuracy in binary classification. We will also be training a classifier from the TPOT library for choosing the best classifier, with respect to accuracy, and also perform hyperparameter tuning on the said classifier, and discovering the best pipeline. We will then evaluate each of the classifiers via k-fold cross-validation and determine the best classifier on the basis of the best average score.

Thus, the final list of the classifier used are:
- SVC
- Perceptron
- KneighborsClassifier
- Decision Tree Classifier
- Naïve Bayes Classifier
- TPOT Classifier

We will limit our discussion to the working of the TPOT classifier, as the other classifiers are standard algorithms, and their implementation have been rigorously documented and discussed in the Scikit-learn Paper (see references).

This is the automation flow of the TPOT library. It is built on top of Scikit-learn and performs hyperparameter tuning on Scikit-learn's regressors and classifiers.

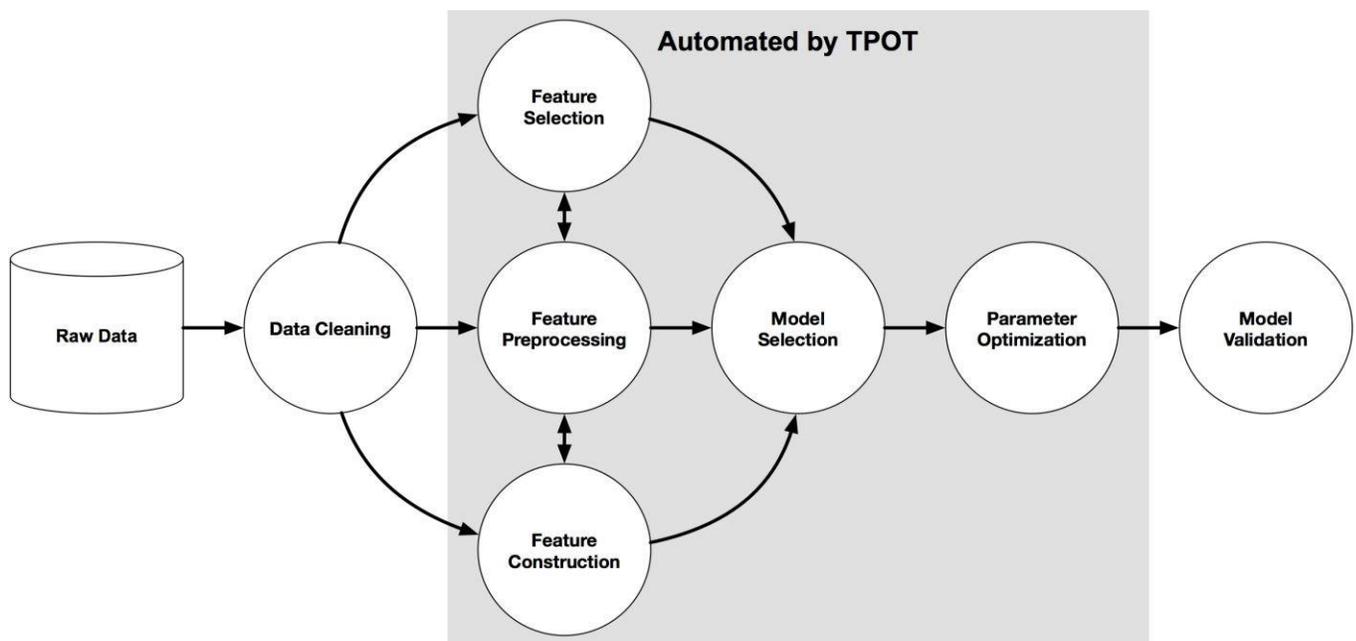

A TPOT classifier will work with thousands of pipelines, and then recommend the pipeline that works the best for the given data. It applies genetic algorithms on an initial population of pipelines to pick the most fit pipelines, reproduce a new generation of pipelines and iterate this process over a number of generations. This process usually converges to a best pipeline, unless it is explicitly terminated by a programmer.

## 3. Results

We obtain the following accuracy scores for the classifiers (note that these output values are taken directly from the python console).

---

1
0.7859531772575251
0.6923076923076923
0.6337792642140468
0.7178631051752922
0.6961602671118531
Average= 0.7052127012132818

```
DecisionTreeClassifier(class_weight=None, criterion='gini', max_depth=None,
            max_features=None, max_leaf_nodes=None,
            min_impurity_decrease=0.0, min_impurity_split=None,
            min_samples_leaf=1, min_samples_split=2,
            min_weight_fraction_leaf=0.0, presort=False, random_state=None,
            splitter='best')
```

2
0.81438127090301
0.7558528428093646
0.7274247491638796
0.7328881469115192
0.7245409015025042
Average= 0.7510175822580555

```
SVC(C=1.0, cache_size=200, class_weight=None, coef0=0.0,
  decision_function_shape='ovr', degree=3, gamma='auto', kernel='rbf',
  max_iter=-1, probability=False, random_state=None, shrinking=True,
  tol=0.001, verbose=False)
```

3
C:\Users\Ritabrata Maiti\Anaconda3\lib\site-packages\sklearn\linear_model\stochastic_gradient.py:128:    FutureWarning: max_iter and tol parameters have been added in <class 'sklearn.linear_model.perceptron.Perceptron'> in 0.19. If both are left unset, they default to max_iter=5 and tol=None. If tol is not None, max_iter defaults to max_iter=1000. From 0.21, default max_iter will be 1000, and default tol will be 1e-3.
  "and default tol will be 1e-3." % type(self), FutureWarning)
0.81438127090301
0.24414715719063546

0.7290969899665551
0.21535893155258765
0.7262103505843072
Average= 0.5458389400394191

Perceptron(alpha=0.0001, class_weight=None, eta0=1.0, fit_intercept=True,
      max_iter=None, n_iter=None, n_jobs=1, penalty=None, random_state=0,
      shuffle=True, tol=None, verbose=0, warm_start=False)

4
0.7474916387959866
0.7408026755852842
0.725752508361204
0.657762938230384
0.7262103505843072
Average= 0.7196040223114333

KNeighborsClassifier(algorithm='auto', leaf_size=30, metric='minkowski',
           metric_params=None, n_jobs=1, n_neighbors=5, p=2,
           weights='uniform')

5
0.81438127090301
0.7558528428093646
0.7290969899665551
0.7813021702838063
0.7262103505843072
Average= 0.7613687249094087

BernoulliNB(alpha=1.0, binarize=0.0, class_prior=None, fit_prior=True)

6
0.8294314381270903
0.7675585284280937
0.7591973244147158
0.7929883138564274
0.7262103505843072
Average= 0.7750771910821269

Pipeline(memory=None,
     steps=[('linearsvc', LinearSVC(C=5.0, class_weight=None, dual=False, fit_intercept=True,
     intercept_scaling=1, loss='squared_hinge', max_iter=1000,
     multi_class='ovr', penalty='l2', random_state=None, tol=0.001,
     verbose=0))])

We observe that the TPOT optimised pipeline has the best average score followed by the Naïve Bayes Classifier. In this case the TPOT optimised pipeline happens to be a LinearSVC classifier with tuned hyperparameters.

(**Note**: We have run the TPOT classifier fit optimisation method for only 5 generations. Better accuracy scores may be achieved by increasing the number of generations)

In conclusion we may note that genetic optimised pipelines work the best in this classification task. However, the Naïve Bayes classifier provides a similar performance, on a k-fold cross- validation test.

## 4. Acknowledgements